\documentclass[letterpaper, 10 pt, conference]{ieeeconf}  % Comment this line out if you need a4paper
\usepackage{graphicx}
\usepackage{caption}

\IEEEoverridecommandlockouts                              % This command is only needed if 
\usepackage{amsmath} % assumes amsmath package installed
\usepackage{booktabs} % for \toprule, \midrule, etc.
\usepackage{float}

\title{\LARGE \bf
Gen2Dex: Learning Dexterous Manipulation from Simulation-Initialized Generated Video Demonstrations
}

\author{Xiaowen Qiu$^{1,*}$, Yian Wang$^{1,*}$, Haoxiang You$^{2}$, and Chuang Gan$^{1}$%
\thanks{Xiaowen Qiu and Yian Wang contributed equally.}%
\thanks{$^{1}$Xiaowen Qiu, Yian Wang, and Chuang Gan are with the University of Massachusetts Amherst, Amherst, MA 01003, USA.}%
\thanks{$^{2}$Haoxiang You is with the Department of Mechanical Engineering, Yale University, New Haven, CT 06520, USA.}%
}

\begin{document}

\IEEEaftertitletext{
    \vspace{-18pt}
    \begin{center}
        \includegraphics[width=\textwidth]{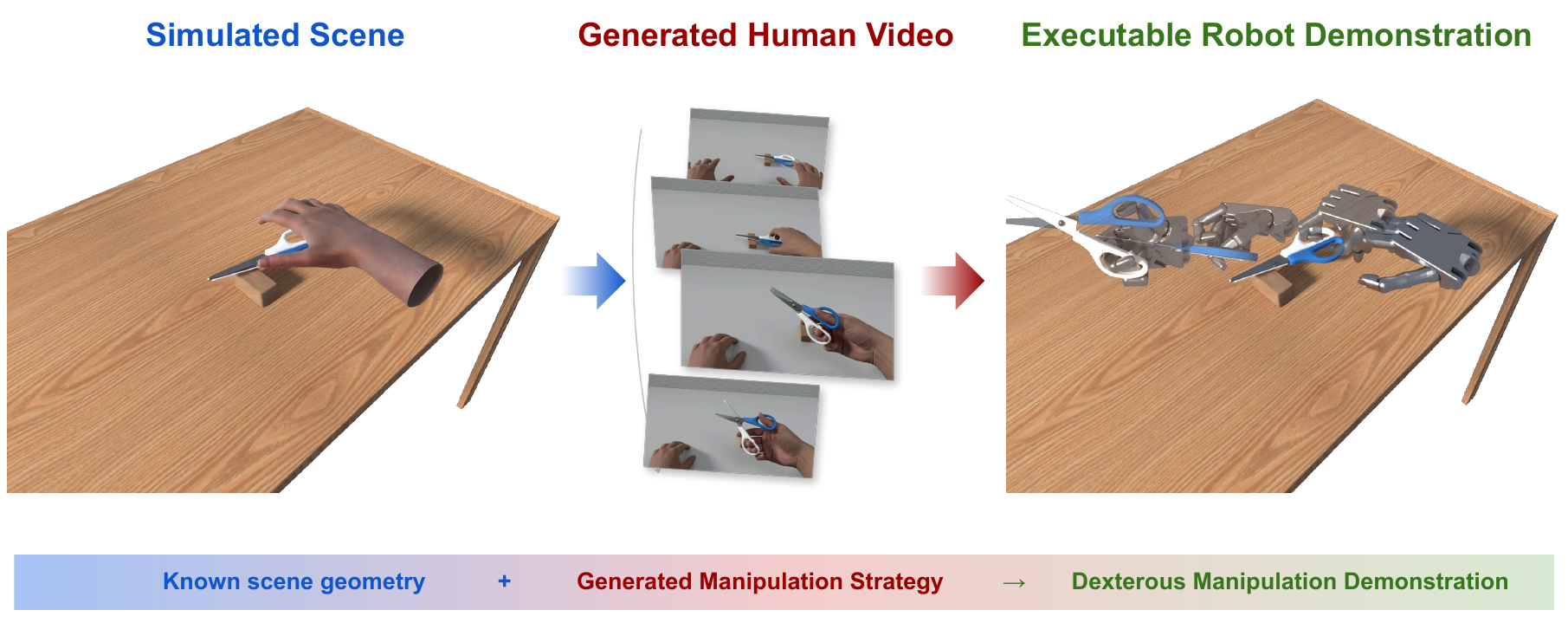}
        \captionof{figure}{Overview of our method.}
        \label{fig:teaser}
    \end{center}
}

\maketitle
\thispagestyle{empty}
\pagestyle{empty}

%%%%%%%%%%%%%%%%%%%%%%%%%%%%%%%%%%%%%%%%%%%%%%%%%%%%%%%%%%%%%%%%%%%%%%%%%%%%%%%%

\begin{abstract}

Human video demonstrations provide rich manipulation strategies for dexterous manipulation learning. We study a setting in which a fully specified simulated scene is used to generate videos that specify the demonstrated manipulation behavior. This setup provides access to scene information such as object geometry, articulation, and camera parameters, while allowing direct control over task and scene configuration. However, generated videos may contain geometric distortions, inaccurate contacts, and other physical inconsistencies, making the recovered motion difficult to execute directly on a robot.

We present an end-to-end framework for converting such simulation-grounded videos into physically executable dexterous robot demonstrations. Our perception pipeline leverages the known scene through model-based object tracking with dense foundation-model features and depth-based hand-object alignment. The recovered human motion is then retargeted and physically grounded using residual reinforcement learning, where a backward curriculum and a cache of physically valid reset states enable the policy to correct imperfect video-derived references while preserving the demonstrated manipulation strategy. On clean ARCTIC demonstrations, our method achieves competitive performance with prior dexterous retargeting approaches and shows greater robustness when applied to references extracted from generated videos. We further demonstrate diverse rigid and articulated manipulation tasks in simulation and successful open-loop transfer of the optimized trajectories to a real dexterous robot.

\end{abstract}

%%%%%%%%%%%%%%%%%%%%%%%%%%%%%%%%%%%%%%%%%%%%%%%%%%%%%%%%%%%%%%%%%%%%%%%%%%%%%%%%

\section{INTRODUCTION}

Recent progress in machine learning has demonstrated the effectiveness of the data-driven paradigm, where scaling diverse training data enables increasingly general models. Robot learning has begun to follow the same trend, with large-scale datasets and generalist policies showing improved transfer across tasks, environments, and embodiments~\cite{o2024open, intelligence2025pi_,generalist2026gen15}. However, collecting robot interaction data remains substantially more expensive than obtaining internet-scale data generated as a natural byproduct of human activity. Robot data require physical hardware, human operators, and repeated interaction with the environment. This data bottleneck is particularly severe for dexterous manipulation, where high-dimensional multi-finger control makes teleoperation and demonstration collection difficult. Human manipulation videos, especially egocentric videos, offer an appealing complementary source of supervision: they can capture rich and diverse manipulation strategies without requiring robot hardware and are available at substantially larger scale~\cite{hoque2026egodex, grauman2022ego4d}. Due to the embodiment gap between human and robot hands, however, such videos cannot be directly treated as robot action trajectories. Earlier works therefore primarily exploited human videos for representation or reward pretraining~\cite{nair2022r3m, chane2023learning}, while recent approaches have begun to explicitly recover and retarget human hand-object interactions into executable dexterous robot motion~\cite{paliwal2026dexterous,wu2026grounding, chen2026dexterous,chen2025vividex, lum2025crossing}.

The annotation available in human demonstrations varies substantially with the data collection setup. At one extreme, datasets collected with specialized sensing systems, such as ARCTIC~\cite{fan2023arctic}, provide accurate 3D hand and object motion, object geometry, and detailed interaction annotations, but require dedicated capture setups and are costly to collect. At the other extreme, monocular egocentric videos are considerably easier to acquire but provide only image observations, with object geometry, metric depth, object motion, and accurate hand-object alignment largely unobserved. Consequently, methods that learn dexterous manipulation from such videos must first recover these quantities through monocular reconstruction and tracking~\cite{paliwal2026dexterous, chen2026dexterous}. While recent advances have made this increasingly feasible, reconstructing both the underlying scene and the manipulation motion from a single video remains inherently ambiguous, and errors in geometry, pose, and hand-object alignment can propagate into downstream retargeting.

Rather than reconstructing both the scene and manipulation motion from an unknown monocular video, we explore a complementary setting that starts from a fully specified simulated scene and uses video generation to synthesize the manipulation behavior. The simulated scene provides object geometry and articulation, initial object poses, and calibrated camera parameters by construction, allowing these quantities to remain available throughout the pipeline. Recent advances in video generation~\cite{seedance2026seedance} further make it possible to synthesize plausible human manipulation conditioned on an initial scene observation~\cite{patel2026roboticmanipulationimitatinggenerated, wu2026grounding}. In this setting, the video generator provides the manipulation strategy, while the underlying simulation provides a controlled and geometrically grounded representation of the scene.

This setting requires both perception to recover hand-object motion and physical grounding to obtain executable robot trajectories. Although the underlying scene geometry and camera parameters are known, generated videos may contain geometric distortions, inaccurate contacts, and locally inconsistent motion. Perception must therefore exploit the known scene while tolerating generation-induced errors. Likewise, physical grounding must treat the recovered motion as an imperfect guide, allowing local deviations for successful manipulation. These challenges motivate our simulation-grounded perception pipeline and reinforcement-learning formulation for refining imperfect references into physically feasible dexterous trajectories.

To address these challenges, we propose an end-to-end framework for learning dexterous manipulation from generated video demonstrations. Given a simulated scene and a generated video, our perception pipeline first recovers the demonstrated object and hand motion. We develop a model-based object tracker that exploits the known object geometry and dense DINOv3~\cite{simeoni2025dinov3} correspondences, together with a depth-mediated hand-object alignment procedure that brings independently reconstructed hand and object trajectories into a consistent metric frame. The recovered human motion is then retargeted to the robot hand and used as a reference for residual reinforcement learning. To cope with imperfect references and the difficult exploration problem of dexterous interaction, we introduce a backward curriculum that progressively learns the task from later to earlier stages of the demonstration, supported by a precomputed cache of physically valid hand-object reset states.

We evaluate our approach both on clean human demonstrations and on references recovered from generated videos. On ARCTIC, our method remains competitive with existing dexterous retargeting approaches when accurate references are available, while showing substantially stronger robustness when applied to imperfect references extracted from generated videos. We further demonstrate the full pipeline on a diverse set of simulated dexterous manipulation tasks, including rigid-object manipulation and articulated-object interaction, and transfer the optimized trajectories to a real robot through open-loop replay. Our main contributions are:
\begin{itemize}
\item We introduce a framework that uses generated human videos to provide manipulation strategies while retaining exact knowledge of the underlying scene.
\item We develop a perception pipeline tailored to this setting, including model-based object tracking with dense vision-foundation-model features and depth-mediated hand-object alignment.
\item We propose a residual RL formulation with a backward curriculum and physically valid reset-state cache, enabling robust trajectory optimization from imperfect video-derived references.
\end{itemize}

\section{RELATED WORK}

\subsection{Dexterous Manipulation from Video Demonstrations}

One line of work leverages large-scale egocentric human videos as pretraining or co-training data, followed by adaptation using comparatively small amounts of robot demonstrations~\cite{kareer2025egomimic,chen2025large,zheng2026egoscale,zhang2026unidex,zhong2026egosteer}. These works show that human motion can provide useful priors for dexterous policies despite the human-robot embodiment gap~\cite{zheng2026egoscale,zhang2026unidex}.

A second line of work uses an individual human video as a task demonstration and converts the demonstrated hand-object motion into robot-executable trajectories through retargeting and physical grounding~\cite{qin2022dexmv,chen2025vividex,paliwal2026dexterous,chen2026dexterous,mu2026deximit,ren2026c2dexcontactconsistentreconstructionretargeting}. Some methods assume or recover relatively accurate hand-object references, while more recent systems additionally reconstruct object geometry, pose, and hand motion directly from monocular RGB video before retargeting~\cite{paliwal2026dexterous,chen2026dexterous,ren2026c2dexcontactconsistentreconstructionretargeting}.

More recently, generated videos have been used as manipulation demonstrations, either by conditioning robot policies directly on generated motion or by extracting explicit trajectories for downstream physical grounding~\cite{bharadhwaj2024gen2act,patel2026roboticmanipulationimitatinggenerated,ye2025textsc,wu2026grounding}. Our work follows the latter direction while starting from a fully specified simulated scene.

\subsection{Hand-Object Reconstruction and Tracking}
A line of work jointly reconstructs hand-object interactions from monocular video, recovering hand motion together with object geometry and pose~\cite{fan2023holdcategoryagnostic3dreconstruction,paliwal2026dexterous,chen2026dexterous,ren2026c2dexcontactconsistentreconstructionretargeting}. Recent video-to-robot systems build on such reconstruction pipelines to obtain temporally consistent hand-object trajectories for downstream retargeting~\cite{paliwal2026dexterous,chen2026dexterous,wu2026grounding}.

Other works focus on individual components of the reconstruction problem. For hands, methods such as HaMeR and HaWoR recover 3D hand pose and motion from monocular or egocentric video~\cite{pavlakos2023reconstructinghands3dtransformers, potamias2025wilorendtoend3dhand,zhang2025haworworldspacehandmotion}. For objects, prior work either jointly reconstructs unknown object geometry and motion from video~\cite{wen2023bundlesdfneural6doftracking,fan2023holdcategoryagnostic3dreconstruction}, or, when an object model is available, directly estimates its pose using model-based methods such as FoundationPose and FoundPose~\cite{wen2024foundationpose,ornek2024foundposeunseenobjectpose}. Our object tracker follows the latter setting, while additionally handling articulated objects and geometric inconsistencies introduced by generated videos.

\subsection{Dexterous-Hand Retargeting and Physical Grounding}
Dexterous-hand retargeting maps human hand motion to robot embodiments with different kinematics. Classical methods primarily optimize geometric hand-pose correspondence~\cite{handa2019dexpilotvisionbasedteleoperation,qin2024anyteleopgeneralvisionbaseddexterous}, while more recent approaches preserve task-relevant hand-object interaction structure, such as contacts or local interaction topology~\cite{wu2026toporetargetinteractionpreservingretargetingdexterous,lai2026objretargetobjectawaremotionretargeting,ren2026c2dexcontactconsistentreconstructionretargeting}.

Because kinematically retargeted trajectories are often not directly executable, another line of work uses physics-based optimization or reinforcement learning to recover physically feasible robot motion~\cite{qin2022dexmv,li2025maniptrans,liu2025dextrackgeneralizableneuraltracking,xu2025dexplorescalableneuralcontrol,feng2026minimalistretargetingguidedreinforcementlearning,wu2026reforcelearningforceawareretargeting}. Some recent methods further incorporate task-specific guidance, including functional object motion in DexMachina, virtual-contact guidance in SPIDER, and contact-wrench guidance in CHORD~\cite{mandi2025dexmachina,pan2025spider, zhu2026learning}. Our method focuses on robustly grounding imperfect video-derived references using residual control together with a backward curriculum and physically valid reset states.

\section{METHOD}

\begin{figure*}[t]
    \centering
    \includegraphics[width=\textwidth,trim=0 54bp 0 0,clip]{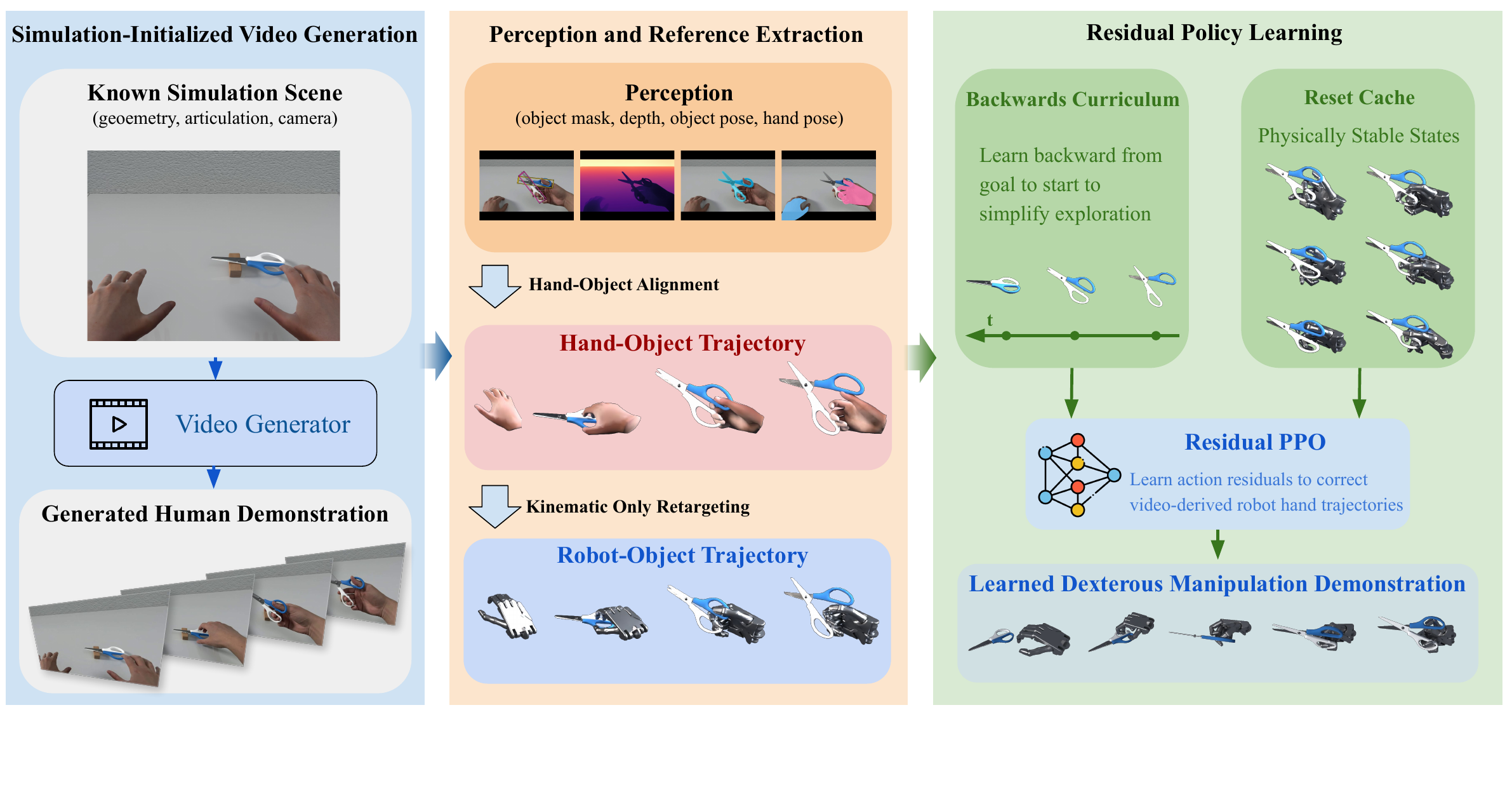}
    \caption{
        Overview of our pipeline for generating physically executable dexterous robot demonstrations from generated videos.
    }
    \label{fig:pipeline}
\end{figure*}
    \vspace{-8pt}

\subsection{Problem Setup}
We assume a known simulated scene with object meshes and poses and calibrated camera intrinsics and extrinsics. Given a task instruction and a video generated from this scene, we seek a robot trajectory that accomplishes the task while following the demonstrated manipulation strategy. The formulation also supports externally provided videos when the required scene information is available.

\subsection{Video Generation}
We place a human hand mesh in the initial simulated scene and use ray tracing to produce a more photorealistic conditioning image, with the aim of improving video generation quality. We condition video generation on this image and a text prompt describing the task. The image provides the initial visual configuration, while the prompt specifies the intended manipulation. We retain the underlying object geometry and camera parameters for subsequent motion recovery, treating the generated video as a behavioral reference rather than a physically accurate simulation.

\subsection{Video-to-Reference Pipeline}
As illustrated in Fig.~\ref{fig:pipeline}, we recover object and hand reference trajectories using SAM 3 segmentation, our optimization-based object tracker, RollingDepth depth estimation, HaWoR hand reconstruction, and our depth-mediated hand-object alignment. The aligned motion is then retargeted to the robot embodiment.
\paragraph{Object tracking}
Generated videos may distort object geometry, making standard learned pose estimators such as FoundationPose less reliable. We also consider articulated objects with changing joint configurations. Given known object geometry and camera calibration, we develop a coarse-to-fine optimization-based tracker inspired by FoundPose~\cite{ornek2024foundposeunseenobjectpose}. We render templates at discrete orientations and joint configurations, retrieve candidates by DINO feature similarity, and establish dense correspondences for the top-ranked templates. The coarse stage applies PnP-RANSAC to rigid objects and minimizes the following objective for articulated objects:
$$\sum_i
\left\|
\mathbf{u}_i-
\pi\left(
\mathbf{T}\mathbf{T}_{\ell_i}(\mathbf{q})\mathbf{X}_i
\right)
\right\|_2^2,
$$
where $\mathbf{X}_i$ is a 3D point on link $\ell_i$, $\mathbf{T}_{\ell_i}(\mathbf{q})$ is its forward-kinematics transform conditioned on joint configuration $\mathbf{q}$, $\mathbf{T}$ is the global root pose, $\mathbf{u}_i$ is the matched image location, and $\pi$ denotes camera projection.

The refinement step further optimizes the coarse pose using the following objective:

$$\sum_i
\left\|
\mathbf{f}_i^T-
F^V\left(\pi(\mathbf{X}_i;\mathbf{T},\mathbf{q})\right)
\right\|_2^2
+
\lambda_{\mathrm{pen}}\mathcal{L}_{\mathrm{pen}}
+
\lambda_{\mathrm{mask}}\mathcal{L}_{\mathrm{mask}},
$$
where $\mathbf f_i^T$ is the DINOv3 descriptor of model/template point $i$ (the superscript $T$ denotes the template), and $F^V$ is the dense DINOv3 feature map of the video frame. The feature term aligns the rendered object with the observed image, while $\mathcal{L}_{\mathrm{pen}}$ and $\mathcal{L}_{\mathrm{mask}}$ penalize configurations with penetration and mask-inconsistent projections.

\paragraph{Hand-Object Alignment}
The reconstructed MANO hands and tracked object are estimated independently and therefore may not share a consistent metric scale or coordinate frame. We use monocular video depth as a common geometric bridge to align them. For each frame $t$, we back-project the predicted depth within the object and hand masks to obtain point clouds $\mathbf P_t^{\mathrm{depth,obj}}$ and $\mathbf P_t^{\mathrm{depth},\mathrm{hand}}$, respectively. We also obtain the metrically tracked object point cloud $\mathbf P_t^{\mathrm{track,obj}}$ and the reconstructed MANO hand point cloud $\mathbf P_t^{\mathrm{MANO},\mathrm{hand}}$. We jointly optimize the depth-to-object and hand-to-depth transformations over the full sequence using gradient-based optimization. Specifically, we minimize the following Chamfer-style alignment objective:

$$
\begin{aligned}
&\mathcal{L}_{\mathrm{align}} = \\
&\sum_t
D_{\mathrm{aug}}
\left(
s\mathbf P_t^{\mathrm{depth,obj}},
\mathbf P_t^{\mathrm{track,obj}}+\mathbf t_t^\mathrm{obj}
\right)
\\
&+
\lambda
\sum_{t,\mathrm{hand}}
D_{\mathrm{aug}}
\left(
s\mathbf P_t^{\mathrm{depth},\mathrm{hand}},
\mathbf P_t^{\mathrm{MANO},\mathrm{hand}}+\mathbf t_t^\mathrm{hand}
\right),
\end{aligned}$$
where $s$ is the global depth scale, $\mathrm{hand}$ indexes the hands, and $\mathbf t_t^{\mathrm{obj}}$ and $\mathbf t_t^{\mathrm{hand}}$ are the per-frame translations aligning the tracked object and each reconstructed hand with the scaled depth geometry, respectively. $D_{\mathrm{aug}}$ is a Chamfer-style distance whose correspondences are selected using both 3D position and DINOv3 features, while the loss itself measures only 3D spatial error. In this way, the tracked object serves as a metric anchor, and the depth map transfers its scale and position to the reconstructed hands.

We use an off-the-shelf retargeter to jointly map the aligned human hand motion to robot wrist poses and finger joint angles. Together with the tracked object trajectory, these kinematic references guide residual RL toward physically executable manipulation.

\subsection{Residual RL Trajectory Optimization}

\paragraph{Residual Policy Learning}
The reference trajectory recovered from video is purely kinematic and may contain tracking errors or physically inconsistent configurations, such as hand-object penetration. We therefore use reinforcement learning to refine this reference into a physically executable trajectory while preserving the demonstrated manipulation strategy. Our policy is trained with PPO and uses a hybrid action space. The wrist follows the retargeted reference trajectory with a bounded residual around the reference wrist pose, allowing the policy to correct errors in the demonstrated wrist motion. The fingers use a delta joint-angle action space independent of the reference finger configuration, giving the policy greater freedom to discover finger motions required for successful interaction. The action space formulated as following:
$$
\begin{aligned}
\mathbf p_t^{w}
&=
\mathbf p_t^{w,\mathrm{ref}}
+
\mathbf S_p \mathbf a_t^{p},
\qquad
\mathbf R_t^{w}
=
\mathbf R_t^{w,\mathrm{ref}}
\circ
\Delta \mathbf R(\mathbf S_r \mathbf a_t^{r}), \\
\mathbf q_t^{f}
&=
\mathbf q_{t-1}^{f}
+
\mathbf S_f \mathbf a_t^{f}
\end{aligned}$$

where $\mathbf p_t^{w}$, $\mathbf R_t^{w}$, and $\mathbf q_t^{f}$ denote the wrist position, wrist orientation, and finger joint-angle targets at time $t$, respectively. $\mathbf p_t^{w,\mathrm{ref}}$ and $\mathbf R_t^{w,\mathrm{ref}}$ are the retargeted reference wrist position and orientation. $\mathbf a_t^{p}$, $\mathbf a_t^{r}$, and $\mathbf a_t^{f}$ are the policy outputs bounded in $[-1,1]$, and $\mathbf S_p$, $\mathbf S_r$, and $\mathbf S_f$ scale the corresponding action ranges. $\Delta \mathbf R(\cdot)$ maps the rotational residual to a rotation, and $\circ$ denotes rotation composition.

The reward is defined as:

$$R = w_\mathrm{obj} r_{\mathrm{obj}} + w_\mathrm{hand} r_{\mathrm{hand}} + w_\mathrm{reg} r_{\mathrm{reg}}$$
The object-tracking term measures errors in object position and orientation, as well as joint configuration for articulated objects. The hand-tracking term measures wrist position and orientation errors and finger joint-angle error relative to the reference. The regularization term discourages excessive and non-smooth actions. This reward encourages the policy to preserve the demonstrated manipulation strategy without requiring exact tracking of the imperfect reference.

\paragraph{Backward Curriculum}

Directly learning long-horizon dexterous manipulation from the beginning of a demonstration is difficult because the hand has a high-dimensional action space, while informative task-level reward signals may depend on first completing a precise sequence of interactions. This issue is particularly severe for imperfect references extracted from generated videos. A clean reference often already brings the hand close to a valid grasp, whereas a noisy reference may contain penetration or follow an infeasible approach path, requiring the policy to deviate from the reference and discover an alternative local path before establishing contact. For example, to successfully track the reference object motion during lifting, the policy may first need to move around a penetrating reference pose, reach a valid grasp configuration, and close the fingers at the correct time. Before a stable grasp is formed, the policy receives little useful feedback from object tracking and is guided primarily by the hand-tracking reward.

Inspired by prior work on reset-based exploration and curriculum learning~\cite{yin2026emergentdexteritydiverseresets}, we train the policy using a backward curriculum. Given the hand and object reference trajectories, we initialize training from physically valid hand-object interaction states near the end of the demonstration, where only a short remaining sequence needs to be explored. Once the policy reliably succeeds from the current starting point, we progressively move the starting point backward along the demonstration until the policy can execute the task from the beginning. This converts a difficult long-horizon exploration problem into a sequence of shorter local exploration problems.

\paragraph{Precomputed Reset-State Cache}

The backward curriculum requires resetting the simulator to physically valid states at different points along the reference trajectory. Directly initializing the robot hand and object from reconstructed kinematic poses is unreliable, especially after contact is established, because the reconstructed poses may contain penetration, unstable contacts, or configurations in which the object immediately falls under gravity. We therefore precompute a cache of physically valid reset states for each reference frame. For each frame, we sample hand configurations near the retargeted reference and candidate contact correspondences between hand surface points and reachable regions on the object surface. We then refine the wrist pose and finger joints using gradient-based optimization by minimizing:
$$\mathcal{L}_{\mathrm{reset}}
=
\lambda_{\mathrm{c}} \mathcal{L}_{\mathrm{contact}}
+
\lambda_{\mathrm{p}} \mathcal{L}_{\mathrm{penetration}}
+
\lambda_{\mathrm{r}} \mathcal{L}_{\mathrm{reg}}$$

where \(\mathcal L_{\mathrm{contact}}\) encourages the selected hand contact points to reach their sampled object-surface targets, \(\mathcal L_{\mathrm{penetration}}\) penalizes hand-object penetration using the object's signed distance field (SDF), and \(\mathcal L_{\mathrm{reg}}\) regularizes the optimized hand configuration toward the sampled reference configuration and valid joint ranges. The resulting configurations are subsequently validated in physics simulation, and only states with negligible penetration and physically stable object behavior are retained. During RL training, the backward curriculum directly resets to these cached states, providing diverse and physically consistent initialization states without repeating the expensive contact optimization and settling procedure online.

\section{Experiments}
\label{sec:experiments}

We design our experiments to answer three questions:
(1) whether our perception pipeline accurately recovers hand-object motion from simulation-initialized generated videos,
(2) whether our residual RL formulation can robustly convert sometimes physically inconsistent demonstrated motions, including implausible contacts and hand-object penetration, into executable robot trajectories, and
(3) whether the complete pipeline can produce successful dexterous manipulation trajectories in simulation and on real hardware.

\subsection{Experimental Setup}

\paragraph{Simulation and Robot Setup}
We conduct all simulation experiments in Genesis using one of three dexterous hands. We use the Shadow Hand for our main simulation experiments and the Wuji Hand for experiments involving real-world deployment, including the corresponding policy training in simulation. We use the Sharpa Hand only for the ARCTIC benchmark to ensure a consistent embodiment when comparing against prior methods. For real-world experiments, we mount the Wuji Hand on an xArm manipulator and execute the trajectories optimized in simulation.

\paragraph{Evaluation Metrics}
For trajectory-tracking evaluations, we use ADD-AUC to compare executed and reference object trajectories, following DexMachina~\cite{mandi2025dexmachina} and CHORD~\cite{zhu2026learning}. References are either annotated ARCTIC trajectories or trajectories reconstructed from generated videos. To evaluate task completion, we use task-specific binary success criteria and report the number of tasks solved. For example, kettlebell lifting is successful when the kettlebell is raised at least 10~cm above the table. The ablation studies use a suite of 10 generated-video tasks, while the end-to-end pipeline comparison uses its six-task rigid-object subset.

\paragraph{Implementation Details}
We train all residual policies with PPO using a batch size of 2,048 unless otherwise specified. All experiments are conducted on a single NVIDIA RTX 4090 GPU.

\subsection{Perception for Simulation-Grounded Generated Videos}
\label{sec:exp_generated_video}

We evaluate our perception pipeline on simulation-grounded generated videos through object-tracking accuracy and downstream manipulation performance.

\paragraph{Comparison Setup}
We compare our model-based object tracker with FoundationPose and evaluate downstream manipulation performance using three combinations of perception and trajectory optimization components from our method and Do-as-I-Do.

\paragraph{Object Tracking}

We compare our object tracker with FoundationPose for 3D object tracking given a known object model. Because generated videos do not provide ground-truth object poses, we evaluate tracking by comparing the object mask rendered from each estimated pose with a reference mask. We obtain the reference masks using SAM3 followed by manual verification, and report both intersection over union (IoU) and recall. IoU captures overall mask agreement but can penalize an otherwise accurate pose when parts of the rendered object are occluded by the hand and therefore absent from the reference mask. Recall complements IoU by measuring how well the estimated pose covers the visible object region. As shown in Table~\ref{tab:object_tracking_quantitative}, our optimization-based tracker achieves higher IoU and recall than FoundationPose, indicating better agreement between projected object geometry and the visible object regions in generated videos.

\begin{table}[t]
\vspace*{4pt}
\centering
\caption{Object-tracking performance on generated videos.}
\label{tab:object_tracking_quantitative}
\begin{tabular}{lcc}
\toprule
Method & FoundationPose  & Ours \\
\midrule
IOU & 0.562 &  \textbf{0.785} \\
Recall & 0.741 &  \textbf{0.953} \\
\bottomrule
\end{tabular}
\end{table}

\paragraph{End-to-end Manipulation Performance}
We compare Do-as-I-Do perception with its model predictive control (MPC), Do-as-I-Do perception with our residual RL, and our perception with our residual RL. All variants use the same six rigid-object tasks from our 10-task suite, since Do-as-I-Do supports only rigid objects.

Table~\ref{tab:endtoend_perception} shows that replacing MPC with our residual RL increases tasks solved from 1/6 to 2/6 with perception fixed. Replacing perception with our pipeline further improves performance to 6/6 with the RL method fixed, supporting the usefulness of our recovered references beyond the optimizer change alone. Together, the perception experiments demonstrate improved image-space object alignment and downstream manipulation success.

\begin{table}[t]
\vspace*{4pt}
\centering
\caption{Perception and trajectory optimization comparison on six rigid-object tasks.}
\label{tab:endtoend_perception}
\begin{tabular}{llc}
\toprule
Perception & Trajectory optimizer & Tasks solved ($\uparrow$) \\
\midrule
Do-as-I-Do & Do-as-I-Do MPC & 1/6 \\
Do-as-I-Do & Our residual RL & 2/6 \\
Ours & Our residual RL & 6/6 \\
\bottomrule
\end{tabular}
\end{table}

\subsection{Robust Trajectory Optimization from Imperfect References}
\label{sec:exp_rl}

\paragraph{Benchmark and Baselines}

Following prior works, we use ARCTIC~\cite{fan2023arctic} dataset as our main quantitative benchmark, as the dataset contain accurate object tracking and hand tracking annotation, and its tasks require dexterity. As for qualitative experiments, we use a suite of our hand crafted simulation environments, paired with video generated using a rendered image of these environments as first frame condition.

We compare against three recent methods for converting hand-object demonstrations into physically feasible dexterous robot motion: \textbf{DexMachina}~\cite{mandi2025dexmachina}, \textbf{SPIDER}~\cite{pan2025spider}, and \textbf{CHORD}~\cite{zhu2026learning}. Their different uses of reference motion and contact information provide a diverse set of baselines for evaluating robustness to clean ARCTIC~\cite{fan2023arctic} annotations and imperfect generated-video references.

All methods use the Sharpa Hand for the ARCTIC sequences and the Shadow Hand for our sequences extracted from generated videos.

\paragraph{Quantitative Results}
We evaluate all baselines and our method on the same seven-task ARCTIC suite used by DexMachina, using the original annotations as references. We additionally evaluate all methods on references extracted from our generated videos using the proposed perception pipeline. We report mean ADD-AUC for both reference sets in Table~\ref{tab:rl_comparison}. While the methods achieve comparable performance on clean references, the performance gap increases substantially on generated-video references, demonstrating the greater robustness of our method to imperfect reference trajectories.

\begin{table}[t]
\centering
\caption{Mean ADD-AUC ($\uparrow$) on clean ARCTIC and generated-video references, averaged over all evaluated sequences in each set.}
\label{tab:rl_comparison}
\setlength{\tabcolsep}{3pt}
\begin{tabular}{lcc}
\toprule
Method & ARCTIC & Generated videos \\
\midrule
DexMachina & 62.40 & 38.20 \\
SPIDER     & 63.51 & 67.06 \\
CHORD      & 56.30 & 41.15 \\
Ours       & \textbf{65.75} & \textbf{76.18} \\
\bottomrule
\end{tabular}
\end{table}

\begin{figure}[t]
    \vspace*{4pt}
    \centering
    \includegraphics[width=\columnwidth]{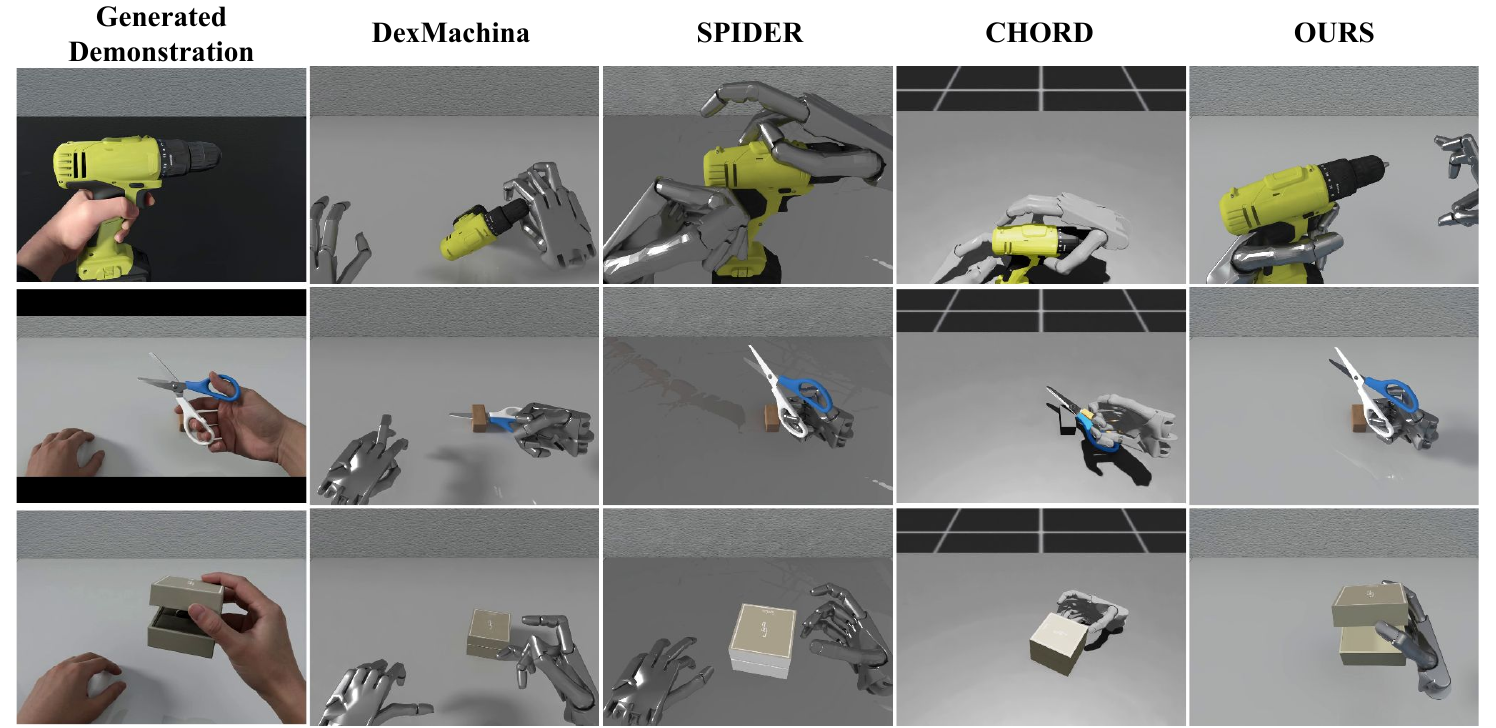}
    \caption{
        Qualitative comparison on our generated-video tasks.
        Each example shows the generated demonstration followed by the best rollout frame from DexMachina, SPIDER, CHORD, and ours.
    }
    \label{fig:sim_compare_exp}
    \vspace{-6pt}
\end{figure}

\paragraph{Qualitative Results}
Figure~\ref{fig:sim_compare_exp} compares generated-video tasks, where our method succeeds consistently while baselines fail on most tasks. In drill handover, the left-hand reference alone is insufficient to grasp the handle: the hand must first open, then close at the appropriate moment. Our backward curriculum exposes the policy to states where the left hand has grasped the handle and the right has released it. Without this curriculum, exploration must discover both grasping and release; in practice, the baselines converge to a suboptimal solution where both hands retain the drill.

\subsection{End-to-End Dexterous Manipulation from Generated Videos}
\label{sec:exp_e2e}
\begin{figure*}[t!]
    \centering
    \includegraphics[width=\textwidth]{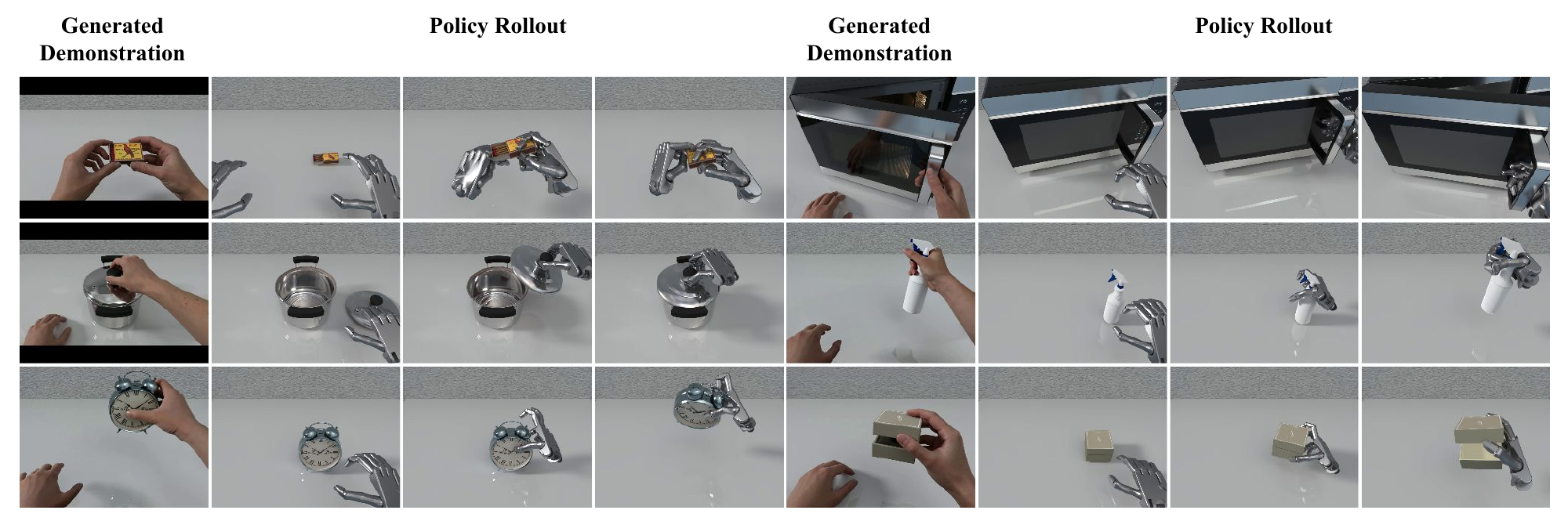}
    \caption{
        Qualitative simulation results. Each example shows a generated video demonstration and keyframes from the corresponding policy rollout.
    }
    \label{fig:sim_Exp}
\end{figure*}
\paragraph{Simulation Results}

Figure~\ref{fig:sim_Exp} shows successful policy rollouts across diverse generated-video tasks. Our method handles both rigid-object manipulation, such as pick-and-place, and articulated-object interactions, such as closing a matchbox and opening a box with one hand.

\paragraph{Real-World Trajectory Replay}

\begin{figure*}[t]
    \centering
    \includegraphics[width=\textwidth]{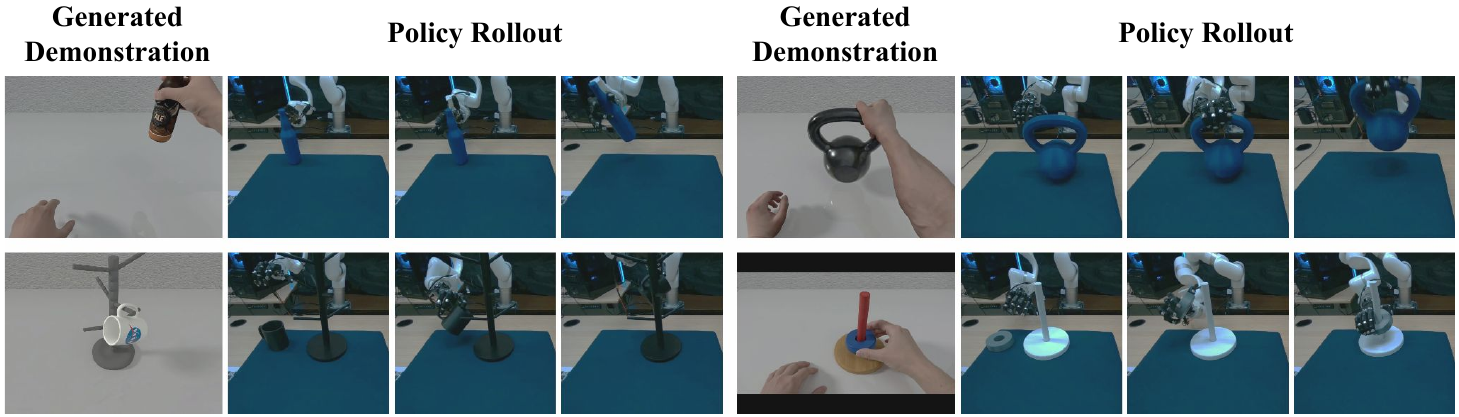}
    \caption{
        Real-world open-loop trajectory replay across multiple dexterous manipulation tasks.
    }
    \label{fig:real_Exp}
    \vspace{-8pt}
\end{figure*}

We transfer the trajectories optimized in simulation to a real xArm equipped with a Wuji Hand and replay them open loop. We evaluate four tasks: pick up bottle, toy-tower insertion, kettlebell lifting, and mug-tree placement. To reproduce the simulated initial conditions, we calibrate the simulated third-person camera to match the real-world setup and manually position each object by aligning it with its rendered 2D silhouette. Figure~\ref{fig:real_Exp} shows that the optimized trajectories transfer successfully to the physical robot across diverse manipulation tasks.

\subsection{Ablation Studies}
\label{sec:exp_ablation}

\paragraph{Reset States and Curriculum}
We ablate reset-state construction and training initialization across 10 tasks. Replacing physically valid reset states with raw reference poses while retaining the backward curriculum yields zero task success; most raw poses contain hand-object penetration and produce unstable resets. We also compare two alternatives without a curriculum: random-frame resets, which sample demonstration frames and use their cached valid states, and start-only RL, which always begins at the task's initial state. Both retain the residual action formulation and rewards. As shown in Table~\ref{tab:rl_ablation}, our method solves 9/10 tasks, compared with 5/10 for random-frame resets and 2/10 for start-only RL. These results support the benefit of progressively extending training from later to earlier demonstration states over either random-frame or first-frame initialization.

\begin{table}[t]
\centering
\caption{Ablation of reset-state construction and training initialization across 10 tasks. Random-frame resets and start-only RL use no curriculum.}
\label{tab:rl_ablation}
\setlength{\tabcolsep}{2.5pt}
\begin{tabular}{lccc}
\toprule
Variant & Reset state & Training start & Tasks solved ($\uparrow$) \\
\midrule
Ours & Valid & Backward & 9/10 \\
Raw-reference resets & Raw & Backward & 0/10 \\
Random-frame resets & Valid & Random frame & 5/10 \\
Start-only RL & Initial & First frame & 2/10 \\
\bottomrule
\end{tabular}
\vspace{-6pt}
\end{table}

\section{CONCLUSIONS}
We presented a framework for converting simulation-initialized generated videos into executable dexterous robot demonstrations. Our perception pipeline improves robustness in this setting through tailored object-tracking and hand-object alignment modules. Residual RL with a backward curriculum and physically valid reset states then refines the recovered references into physically executable trajectories. Experiments demonstrate competitive performance on clean ARCTIC references and greater robustness on generated-video references. We further demonstrate diverse rigid and articulated manipulation tasks in simulation and successful open-loop trajectory transfer to a real robot.


%%%%%%%%%%%%%%%%%%%%%%%%%%%%%%%%%%%%%%%%%%%%%%%%%%%%%%%%%%%%%%%%%%%%%%%%%%%%%%%%

%%%%%%%%%%%%%%%%%%%%%%%%%%%%%%%%%%%%%%%%%%%%%%%%%%%%%%%%%%%%%%%%%%%%%%%%%%%%%%%%

\bibliographystyle{IEEEtran}
\bibliography{references}

\end{document}